\pdfoutput=1

\documentclass[11pt]{article}

\usepackage[final]{acl}

\usepackage{times}
\usepackage{latexsym}
\usepackage[T1]{fontenc}
\usepackage[utf8]{inputenc}
\usepackage{microtype}

\usepackage{algorithm}
\usepackage{algpseudocode}
\usepackage{graphicx}
\usepackage{array}
\usepackage{svg}
\usepackage{amsmath}
\usepackage{float}
\usepackage{caption}
\usepackage{subcaption}
\usepackage{soul}
\usepackage{multirow}
\usepackage{multicol}
\usepackage{numprint}
\usepackage{longtable}
\usepackage{makecell}
\usepackage{tikz}
\usepackage{booktabs, multirow} % for borders and merged ranges
\usepackage{changepage, threeparttable} % for wide tables
\usepackage{xcolor}
\usepackage{enumitem}
\usepackage{lipsum}
\usepackage{amssymb}
\usepackage{tabularx}
\usepackage{rotating}
\usepackage{geometry}
\usepackage{pifont}
\usepackage{listings}  % For source code listing
\usepackage{xcolor}  % For defining and using colors
\usepackage{textcomp}
 
\newcommand{\mytexttilde}{\raisebox{0.5ex}{\texttildelow}}

\definecolor{codered}{HTML}{FF0000}
\definecolor{codeblue}{HTML}{0000FF}

\newcolumntype{H}{>{\setbox0=\hbox\bgroup}c<{\egroup}@{}}

\graphicspath{{../}}

\newcommand{\cmark}{\text{\ding{52}}}
\newcommand{\xmark}{\text{\ding{56}}}

\lstdefinestyle{json}{
    basicstyle=\ttfamily,
    stringstyle=\color{codered},
    keywordstyle=\color{codeblue}
}

\definecolor{GoodGreen}{RGB}{3, 146, 94}
\definecolor{BadRed}{RGB}{250, 70, 70}

\newcommand{\rot}[1]{\rotatebox{90}{\parbox{2cm}{\centering #1}}}

\usepackage[capitalize,noabbrev]{cleveref}
\crefname{ExNo}{Example}{Examples}
\crefname{SubExNo}{Example}{Examples}
\creflabelformat{SubExNo}{(#2#1#3)}
\creflabelformat{ExNo}{(#2#1#3)}
\crefrangelabelformat{SubExNo}{(#3#1#4--#5\crefstripprefix{#1}{#2}#6)}
\crefrangelabelformat{ExNo}{(#3#1#4)--(#5#2#6)}
\crefname{chapter}{Chapter}{Chapters}
\crefname{section}{\S}{\S\S}
\Crefname{section}{\S}{\S\S}
\crefname{table}{Table}{Tables}
\crefname{figure}{Figure}{Figures}
\crefname{algorithm}{Algorithm}{}
\crefname{equation}{Eq.}{}
\crefname{appendix}{Appendix}{}
\crefformat{section}{\S#2#1#3}
\crefname{lemma}{Lemma}{Lemmas}
\crefname{proposition}{Proposition}{Propositions}
\crefname{definition}{Definition}{Definitions}
\crefname{corollary}{Corollary}{Corollaries}
\crefname{example}{Example}{Examples}
\crefname{problem}{Problem}{Problems}
\crefname{remark}{Remark}{Remarks}

\usepackage{todonotes}
\newcommand{\note}[4][]{{\todo[author=#2,color=#3,size=\footnotesize,fancyline,caption={},#1]{#4}}}
\definecolor{tticblue}{RGB}{0, 94, 184}
\newcommand{\freda}[2][]{{\note[#1]{freda}{tticblue!20}{#2}}}
\newcommand{\Freda}[2][]{\freda[inline,#1]{#2}\noindent}
\newcommand{\mike}[2][]{{\note[#1]{mike}{pink}{#2}}}
\newcommand{\Mike}[2][]{\mike[inline,#1]{#2}\noindent}

\title{
    SpaRE: Enhancing Spatial Reasoning in Vision-Language Models \\
    with Synthetic Data
}

\author{
  Michael Ogezi \\
  University of Waterloo \\
  Vector Institute \\
  \href{mailto:mogezi@uwaterloo.ca}{\color{black}\texttt{mogezi@uwaterloo.ca}} \And 
  Freda Shi \\
  University of Waterloo \\
  Vector Institute \\
  \href{mailto:mogezi@uwaterloo.ca}{\color{black}\texttt{fhs@uwaterloo.ca}} 
}

\begin{document}
\maketitle

\begin{abstract}
    Vision-language models (VLMs) work well in tasks ranging from image captioning to visual question answering (VQA), yet they struggle with spatial reasoning, a key skill for understanding our physical world that humans excel at.
    % We find that while a few most frequent spatial relations are reasonably well represented, the remaining 95\%, which form a long tail of rare spatial relations, appear in only 0.02\% of samples in widely used VL datasets.
    We find that spatial relations are generally rare in widely used VL datasets, with only a few being well represented while most form a long tail of underrepresented relations.
    This gap leaves VLMs ill-equipped to handle diverse spatial relationships.
    To bridge it, we construct a synthetic VQA dataset focused on spatial reasoning generated from hyper-detailed image descriptions in Localized Narratives, DOCCI, and PixMo-Cap.
    Our dataset consists of 455k samples containing 3.4 million QA pairs.
    Trained on this dataset, our {Spa}tial-{R}easoning {E}nhanced (SpaRE)
    % \freda{We'd a full name of SpaRE here.} 
    VLMs show strong improvements on spatial reasoning benchmarks, achieving up to a 49\% performance gain on the What's Up benchmark, while maintaining strong results on general tasks.
    Our work narrows the gap between human and VLM spatial reasoning and makes VLMs more capable in real-world tasks such as robotics and navigation.
    We look to share our code and dataset in due course.
\end{abstract}

\section{Introduction}
\label{sec:introduction}

\begin{figure}[t]
    \centering
    \includegraphics[width=\linewidth]{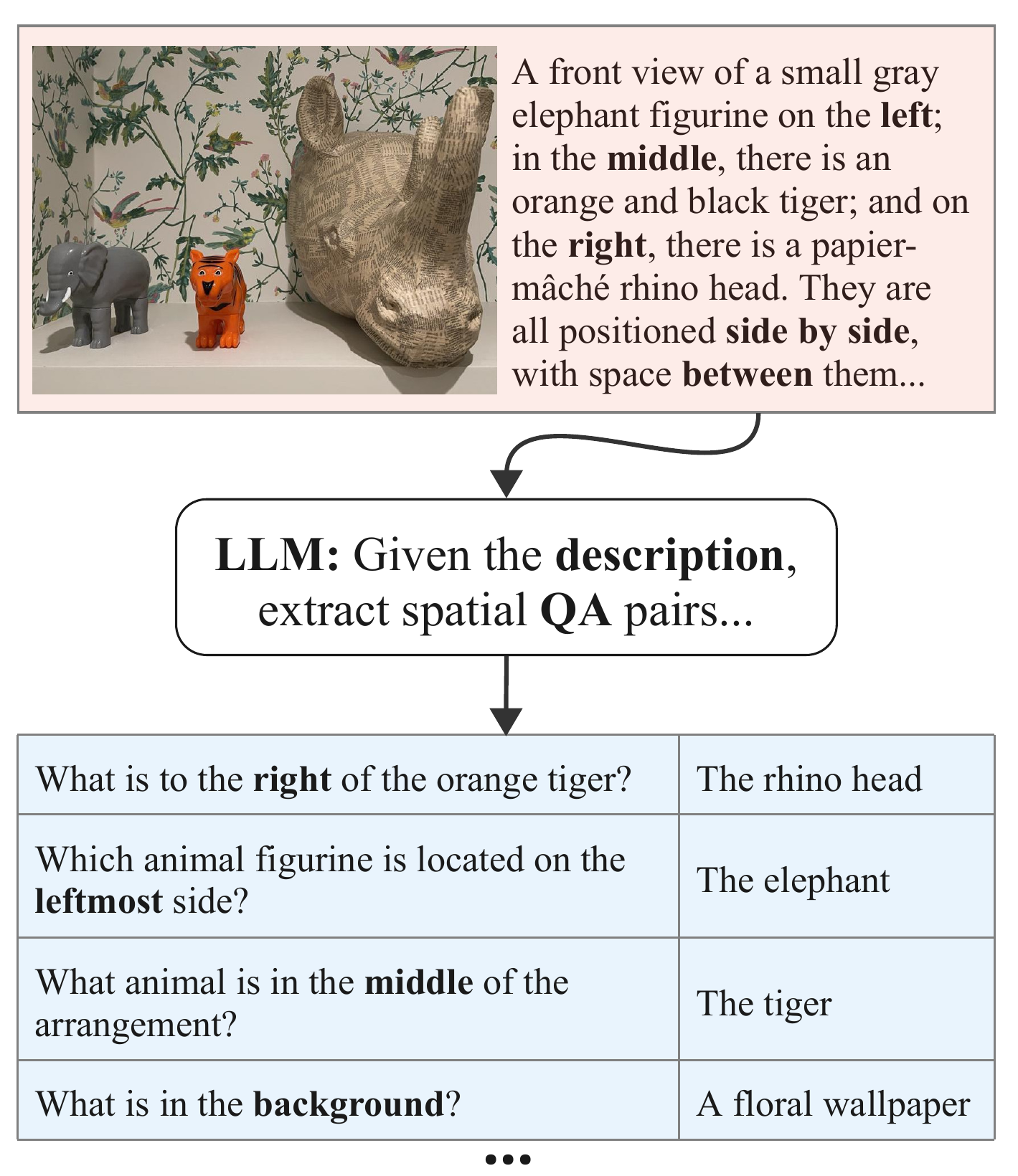}
    \caption{
        Our synthetic data generation pipeline: Hyper-detailed image descriptions are fed to an LLM that extracts spatial-reasoning question-answer (QA) pairs.
    }
    \label{fig:method_summary}
\end{figure}

% \Freda{The intro tells the story very well! I think the only things we need to work on about it are (1) a figure to illustrate the method, and (2) shorten it a little bit.}
% \Mike{
%    1. I've added the figure
%    2. It's shorter but I could do a bit more.
%}

Spatial reasoning—the ability to understand and interpret spatial relationships between objects—is a critical component of intelligent systems that interact with the physical world \citep{Newcombe2000MakingST}. 
Applications such as robotics, autonomous navigation, and extended reality rely heavily on precise spatial understanding to function effectively \citep{Landsiedel2017A, Balakrishnan2021InteractionOS}. 
Without robust spatial reasoning, these systems risk misinterpreting their environments, leading to failures that could compromise safety and efficiency.

% \freda{Quite a minor thing here: I don't think the \texttt{cite} command needs a tilde before it; also I'm more used to \texttt{citep} for easier transition across template formats. }

%\mike{Tilde stops line breaks btwn citation and predecessor word}
%\freda{yes I mean... it's okay to break :)}

Despite impressive advancements in vision-language models (VLMs) for tasks like image captioning, visual question answering (VQA), image-text retrieval, and zero-shot image classification, these models consistently struggle with spatial reasoning~\citep{kamath-etal-2023-whats, liu-etal-2023-visual, zhang2024visionlanguagemodelsrepresentspace}. For instance, VLMs may correctly identify objects in an image but fail to comprehend their spatial arrangement, which is crucial for tasks like scene understanding and navigation.

% \freda{I feel this paragraph could be combined with the next one, and to be shortened a little bit.  Some work could be cited in related work. }
% \mike{Done. Condensed by about 50\%}

Spatial relations are rare in existing vision-language datasets (see Table~\ref{tab:dataset_analysis}). 
Common relations (like on, left, and under) dominate, while less frequent ones (like facing, opposite, and surrounding) are severely underrepresented. 
In fact, the top 17\% of relations make up over 90\% of all spatial relation examples (see Table~\ref{tab:basic_spatial_relations_taxonomy} in the appendix). 
This imbalance leaves VLMs poorly equipped to handle the full range of spatial relations.

Previous efforts to address this gap have fallen short.
Synthetic datasets \citep{johnson2017clevr, agrawal2023stupd}, while providing structured and controlled environments, rely on simplistic geometric shapes and synthetically generated images, which fail to generalize to real-world data.
On the other hand, human-curated datasets \citep{liu-etal-2023-visual, kamath-etal-2023-whats} are limited in both quantity and diversity of spatial relations, leading to continued subpar performance.

To bridge this gap, we present a novel approach that leverages the untapped potential of hyper-detailed captions from recent long-form image-captioning datasets. Datasets such as DOCCI~\citep{OnoeDocci2024}, PixMo-Cap~\citep{deitke2024molmopixmoopenweights}, and Localized Narratives~\citep{PontTuset_eccv2020} contain rich, detailed descriptions of images, often including explicit descriptions of spatial relationships and object interactions (see Figure~\ref{hyper_detailed_captions} in the appendix).

Armed with these resources, we use \texttt{Qwen2.5-3B-Instruct}~\cite{yang2024qwen2}\footnote{\url{https://hf.co/Qwen/Qwen2.5-3B-Instruct}} to extract synthetic question-answer (QA) pairs focused on spatial relationships. Specifically, we extract spatial information from the hyper-detailed captions and formulate diverse and complex QA pairs that probe various aspects of spatial reasoning. Figure~\ref{fig:method_summary} illustrates our synthetic dataset generation method. By maintaining visual realism through the use of real-world images, our approach allows models to learn spatial reasoning skills in contexts they will encounter in practical applications, effectively addressing the domain gap introduced by synthetic visuals.
 
We fine-tune VLMs on our synthesized dataset to produce \textbf{Spa}tial-\textbf{R}easoning \textbf{E}nhanced (\textbf{SpaRE}) VLMs.
SpaRE VLMs significantly improve performance on spatial reasoning benchmarks, including VSR, What's Up, 3DSRBench, and RealWorldQA. 
As shown in Table~\ref{tab:main_results}, we achieve up to a {49\%} gain on the A split of What's Up~\citep{kamath-etal-2023-whats}, a benchmark designed to test spatial understanding. 
Importantly, these enhancements do not come at the expense of general VL performance.
SpaRE models maintain their performance on standard benchmarks such as MMMU~\cite{yue2024mmmu},
% \freda{Did you mean MMMU? Also, we probably need a citation here.}
and MMBench~\citep{liu2024mmbench}. 
This demonstrates that our method enhances spatial reasoning capabilities while preserving overall model effectiveness.

% \begin{table}[ht]
% \small
% \centering
% \begin{tabular}{@{}llll@{}}
% \toprule
%  & \textbf{Datasets} & \textbf{Total} & \textbf{Percentage} \\
% \midrule
% \multirow{5}{*}{{Captioning}} & TextCaps & 109,765 & 0.248\% \\
%  & ShareGPT-4o & 50,000 & 0.113\% \\
%  & InternVL-SA-1B-Caption & 12,354,191 & 27.93\% \\
%  & NewYorkerCaptionContest & 385 & 0.00087\% \\
%  & MMInstruct & 973,000 & 2.20\% \\
% \midrule
% \multirow{8}{*}{{VQA}} & VQAv2 & 443,757 & 1.003\% \\
%  & GQA & 943,000 & 2.131\% \\
%  & OKVQA & 9,009 & 0.02036\% \\
%  & Visual7W & 327,929 & 0.741\% \\
%  & VSR & 7,680 & 0.01736\% \\
%  & FSC147 & 6,135 & 0.01385\% \\
%  & Objects365-YorN & 29,000,000 & 65.51\% \\
%  & Hateful-Memes & 10,000 & 0.0226\% \\
% \midrule
%  & \textbf{Total} & \textbf{44,234,851} & \textbf{100\%} \\
% \bottomrule
% \end{tabular}
% \caption{Captioning and Question Answering Dataset Details with Percentages}
% \end{table}

By enhancing spatial reasoning in VLMs, our work supports systems that rely on accurate spatial understanding. 
This includes applications such as self-driving cars handling complex roads, robots operating alongside humans, and assistive technologies aiding visually impaired individuals in navigation.

In summary, our contributions are threefold:

% \freda{Just a note in case we need more space: \texttt{enumitem} (a package) gives better indentation.}
\begin{enumerate}[leftmargin=*]
    \setlength{\itemsep}{2pt}
    
    \item \textbf{Quantifying Data Scarcity}: We analyze spatial relations in current datasets and find that the top 17\% account for about 90\% of the samples, revealing a significant gap in representation.
    
    \item \textbf{Synthetic Spatial Data Generation}: We develop a method to generate synthetic spatial reasoning QA pairs from hyper-detailed captions of over one million real-world images using advanced LLMs.
    
    \item \textbf{Enhancing VLM Spatial Reasoning}: Our approach significantly improves VLMs' spatial reasoning capabilities---by up to 49\%---without compromising general VL task performance.
\end{enumerate}

The rest of the paper is organized as follows. In Section~\ref{sec:related_work}, we review related work in spatial reasoning and VLMs. Section~\ref{sec:methodology} details our approach to generating synthetic QA pairs and augmenting training data. We present our experiments and results, and a discussion in Section~\ref{sec:experiments_results}.
Finally, we conclude and briefly outline possible future research directions in Section~\ref{sec:conclusion}.

\section{Background and Related Work}
\label{sec:related_work}

% Spatial Reasoning in VLMs
\subsection{Spatial Reasoning Abilities in VLMs}

Spatial reasoning remains a challenge for VLMs.
\citet{liu-etal-2023-visual} introduced VSR, a dataset consisting of image-caption pairs where the binary task is to predict whether the caption accurately describes the spatial relations observed in the image.
Their findings demonstrate that models consistently underperform on these tasks.
Similarly, \citet{kamath-etal-2023-whats} highlighted that state-of-the-art VLMs struggle with basic spatial relations, performing near random on benchmarks designed to test understanding of concepts like \textit{left}, \textit{right}, \textit{above}, and \textit{below}.
Furthermore, both \citet{gokhale2023benchmarkingspatialrelationshipstexttoimage} and \citet{Cho2022DALLEVALPT} show that text-to-image generation models also struggle with producing images that faithfully represent spatial relations between multiple objects.
While \citet{chen2024spatialvlm} advanced quantitative spatial reasoning through their data generation pipeline (i.e., predicting approximate distances between objects), qualitative spatial understanding remained unexplored.
These findings indicate a significant gap in current models' abilities to process and reason about spatial information.

% Datasets for Spatial Reasoning
\subsection{Datasets for Spatial Reasoning}

% Our analysis of VLM supervised fine-tuning datasets as well as past work by \citet{kamath-etal-2023-whats} and \citet{liu-etal-2023-visual} shows that the scarcity of data samples rich in spatial information is a major barrier to improving spatial reasoning in VLMs.
% \freda{This sentence gives me a pause here. We can probably start with the intro of datasets, and put this sentence into our contribution }
% \mike{moved in a condensed for to contributions in context}

\paragraph{Natural Data}
Popular high-quality, supervised-fine-tuning datasets which we analyze, such as TextCaps~\citep{sidorov2020textcaps}, ShareGPT-4o~\footnote{\url{https://sharegpt4o.github.io/}}, InternVL-SA-1B-Caption~\footnote{\url{https://hf.co/datasets/OpenGVLab/InternVL-SA-1B-Caption}}, NewYorkerCaptionContest~\footnote{\url{https://hf.co/datasets/jmhessel/newyorker_caption_contest}}, MMInstruct~\citep{liu2024mminstruct}, VQAv2~\citep{goyal2017making}, GQA~\citep{Hudson_2019_CVPR}, OKVQA~\citep{marino2019ok}, Visual7W, FSC147~\citep{ranjan2021learning}, Objects365-YorN~\citep{Shao_2019_ICCV}, and Hateful-Memes~\citep{kiela2020hateful} lack enough samples that probe spatial knowledge.
They predominantly focus on object recognition, captioning, and general-purpose VQA, without detailed spatial annotations, leading to a significant data deficiency.

Efforts to tackle this issue include natural datasets such as VSR~\citep{liu-etal-2023-visual}, and What's Up~\citep{kamath-etal-2023-whats} which manually curate training data specifically targeted at spatial reasoning.
However, these datasets are typically small (e.g., the aforementioned ones total around 8k samples).

\paragraph{Synthetic Data}
Synthetic data has been employed to augment training datasets for various language models~\citep{gunasekar2023textbooks, li2023textbooks, abdin2024phi}.
The same has been the case for VLMs in tasks such as image captioning and text-to-image generation~\citep{BetkerImprovingIG}, as well as VQA and general visual instruction tuning~\citep{liu2024visual, chen2025sharegpt4v}.
In the specific case of spatial reasoning, datasets like CLEVR~\citep{johnson2017clevr} and STUPD~\citep{agrawal2023stupd} propose learning from images rendered from 3D synthetic environments with controlled spatial relationships.
Unfortunately, these fail to capture the complexity, and nuance found in natural, real-world images.
As a result, those models suffer from domain-shift issues and achieve poor generalization to practical applications~\citep{agrawal2023stupd}.

\paragraph{Hyper-Detailed Image Descriptions}
Recently, work toward curating datasets that describe images in extreme detail to address the shortcomings~\citep{BetkerImprovingIG} of basic descriptions pulled from alt texts.
Efforts such as DOCCI~\citep{OnoeDocci2024}, PixMo-Cap~\citep{deitke2024molmopixmoopenweights}, and to a lesser extent, Localized Narratives~\citep{PontTuset_eccv2020} which total around 1 million image-description pairs, provide rich visual descriptions of natural images.
In these rich descriptions, we find detailed descriptions of the spatial relationships between objects in the images.
We show an example from DOCCI in Figure~\ref{hyper_detailed_captions} in the appendix.
We look to leverage these datasets to produce synthetic QA data in a manner similar to \citep{liu2024visual, chen2025sharegpt4v}.

% Our Contributions in Context
\subsection{Our Contributions in Context} 
% \freda{I like this way of organizing the related work!}
% \mike{Thanks, I thought it was useful for some re-grounding}
In summary, existing approaches to improving the spatial reasoning abilities of VLMs fall short in terms of diversity, performance, dataset size, and generalization.
We quantify the data scarcity problem and leverage hyper-detailed captions to synthetically generate QA pairs that probe spatial reasoning in a manner that mirrors real-world complexities.

\section{Method}
\label{sec:methodology}

Our objective is to enhance the spatial reasoning capabilities of VLMs with our approach of generating synthetic QA pairs from hyper-detailed image descriptions using an LLM.
In this section, we provide a comprehensive description of this approach.

% \begin{figure*}[t]
%     \centering
%     \includegraphics[width=\linewidth]{../images/proposed_approach.jpg}
%     \caption{Synthetic data generation pipeline.}
%     \label{fig:method}
% \end{figure*}

\subsection{Data Sources and Analysis}

\subsubsection{Selection of Hyper-Detailed Datasets}

To generate a substantial amount of spatial reasoning data, we selected the following three hyper-detailed image-description datasets: DOCCI~\citep{OnoeDocci2024}, Localized Narratives~\citep{PontTuset_eccv2020}, and PixMo-Cap~\citep{deitke2024molmopixmoopenweights}.
We show dataset statistics in Table~\ref{tab:dataset_summary}.

\begin{table}[ht]
    \centering
    \small
    \setlength{\tabcolsep}{4pt}
    \begin{tabular}{lrrr r}
        \toprule
        \textbf{Source}   &
        \textbf{Size}     &
        \textbf{Filtered} &
        \textbf{Words}    &
        \textbf{Gen. Pairs}                         \\
        \midrule
        % 10k, 260k, 276k = 547k filtered
        % 10k, 232k, 214k = 455k finally retrieved
        DOCCI             & 15k    & 10k  & 136 & 108k \\
        LN                & 849k   & 232k & 42  & 1,226k \\
        Pixmo-Cap         & 717k   & 214k & 196 & 2,038k \\
        \midrule
        \textbf{Total}    & 1,581k & 455k & 113 & 3,372k \\
        \bottomrule
    \end{tabular}
    \caption{Details of selected image-description datasets. LN refers to Localized Narratives and Gen. Pairs refers the the number of generated QA pairs.}
    \label{tab:dataset_summary}
\end{table}

\paragraph{DOCCI} 
% \freda{I think we could have \texttt{noindent + textbf} here to save space}
% \mike{Pausing for now since \texttt{noindent + textbf} makes it a bit harder to read and we don't have serious space issues}
~\citep{OnoeDocci2024} features long, human-annotated English descriptions originally-curated images, designed to address challenges such as spatial relations and world knowledge, with each description providing fine-grained detail for improved model training.
We show an example in Figure~\ref{hyper_detailed_captions} in the appendix.

\paragraph{Localized Narratives}
~\citep{PontTuset_eccv2020} offers a unique form of multi-modal image annotation by synchronizing spoken descriptions (which are transcribed) with mouse traces across images from COCO~\citep{lin2015microsoftcococommonobjects}, Flickr30k~\citep{young-etal-2014-image-flickr30k}, ADE20K~\citep{zhou2019semantic} and Open Images~\citep{kuznetsova2020open}, to enhance applications such as controlled image captioning.
We show an example in Figure~\ref{hyper_detailed_captions_ln} in the appendix.

\paragraph{PixMo-Cap}
~\citep{deitke2024molmopixmoopenweights} is a high-quality pre-training dataset featuring a diverse array of images paired with detailed, dense captions created by transcribing and refining spoken descriptions from annotators across approximately 70 topics, to provide rich contextual information for model training.
We show an example in Figure~\ref{hyper_detailed_captions_pixmo} in the appendix.

\subsubsection{Analysis of Spatial Relation Presence}
\label{subsec:dataset_analysis}

To quantify spatial reasoning data in existing VLM datasets, we prompt \texttt{Qwen2.5-3B-Instruct}\footnote{\url{https://hf.co/Qwen/Qwen2.5-3B-Instruct}}~\citep{yang2024qwen2} to identify spatial relations in given descriptions. 
This method performs well, and the prompt used is shown in Table~\ref{tab:spatial_relation_check_prompt} in the appendix.

We applied this approach to several popular VLM datasets, as shown in Table~\ref{tab:dataset_analysis}, where the model predicted the presence of spatial relations in each caption. 
We then matched keywords in the text to compute frequencies for various spatial relations. 
The full statistics are provided in Table~\ref{tab:basic_spatial_relations_taxonomy} in the appendix.

\begin{table}[t]
    \small
    \centering
    \begin{tabular}{@{}llrr@{}}
        \toprule
         & \textbf{Datasets}          & \textbf{Total}              & \textbf{\%}                 \\
        \midrule
        \multirow{8}{*}{\rot{VQA}}
         & VQAv2                      & 443.8k                      & 1.44                        \\
         & GQA                        & 943.0k                      & 3.07                        \\
         & OKVQA                      & 9.0k                        & 0.03                        \\
         & Visual7W                   & 327.9k                      & 1.07                        \\
         & \textbf{\color{blue}{VSR}} & \textbf{\color{blue}{7.7k}}   & \textbf{\color{blue}{0.03}}   \\
         & FSC147                     & 6.1k                        & 0.02                        \\
         & Objects365-YorN            & 29,000.0k                   & 94.35                       \\
         & Hateful-Memes              & 10.0k                       & 0.03                        \\
        \midrule \midrule
         &                            & \textbf{30,747.5k}          & \textbf{100}                \\
        \bottomrule
    \end{tabular}
    \caption{
    VQA datasets in the supervised fine-tuning set used by \texttt{InternVL2}~\citep{chen2024farinternvl1_5}, a leading open-source VLM family.
    The spatial reasoning datasets are in {\color{blue}{\textbf{blue}}}.
    }
    \label{tab:dataset_analysis}
\end{table}

\subsection{Synthetic Data Generation}

\subsubsection{Generation Pipeline}

We use \texttt{Qwen2.5-3B-Instruct}~\citep{yang2024qwen2}\footnote{\url{https://hf.co/Qwen/Qwen2.5-3B-Instruct}} to generate QA pairs focused on spatial reasoning from the image descriptions.
The generation process involves the following:

\begin{enumerate}[leftmargin=*]
    \item \textbf{Pre-Filtering}
          From each dataset, we filter for only descriptions containing explicit spatial information.
          We employ a setup similar to that described in our dataset analysis in Subsection~\ref{subsec:dataset_analysis} to classify viable descriptions.
          Filtering in this way trims our combined datasets by \textasciitilde 65\%, as we show in Table~\ref{tab:dataset_summary}.

    \item \textbf{Prompt Construction and QA Pair Generation}
          We construct a detailed prompt to guide the LLM in extracting relevant and diverse QA pairs, restricted to spatial reasoning.
          During generation, we decode with a temperature of $0$, generating up to a maximum of $8,192$ new tokens.
          We also enforce the generation of structured output in the form of a JSON list of QA pairs for easy parsing.
          The generated pairs are guided to cover positions, orientations, and distances while excluding non-spatial details.
          The full prompt is shown in Table \ref{tab:spatial_relation_qa_generation_prompt} in the appendix.

    \item \textbf{Post-Generation Quality Assurance}
          To ensure that we produce high-quality QA pairs that are relevant to spatial knowledge in the final dataset, we apply a set of automated verification techniques and drop pairs that fail them.
          We discuss these techniques in more detail in Subsection~\ref{subsec:quality_assurance}.
\end{enumerate}

\subsubsection{Dataset Composition}

By applying this method across the selected datasets, we generated a substantial synthetic dataset of spatial reasoning QA pairs. Table~\ref{tab:dataset_summary} summarizes the number of QA pairs generated from each dataset.
% Table~\ref{fig:qa_gen} provides examples of generated QA pairs from a description corresponding to the show image.
Figure~\ref{fig:method_summary} provides examples of generated QA pairs from a description corresponding to the show image.

\subsection{Quality Assurance}
\label{subsec:quality_assurance}

To ensure the quality and accuracy of the generated QA pairs, we implemented automated quality assurance measures.
We employ the following criteria:

\begin{enumerate}[leftmargin=*]
    \item \textbf{Deduplication}
          We check for duplicates among the set of QA pairs generated for each sample and remove them.
          Specifically, we employ full-string matching on the questions.
          and CLIP~\citep{radford2021learning} semantic similarity with a cutoff of $0.95$ which we selected by manually testing a sample of QA pairs from 25 sample images.

    \item \textbf{Reference Check}
        We filter out samples that make references to the description instead of directly asking about the image by matching for keywords such as ``mention,'' and ``description.''

    \item \textbf{Answer-Description Consistency Check}
          We check that the answers are present in the original description to maximize groundedness.
          Specifically, we verify that subsets of the answer, such as key phrases, are present in the description, even if the entire answer is not matched exactly.

    \item \textbf{Image-Question Consistency Check}
          We compare the semantic similarity between images and questions in QA pairs to gauge alignment.
          Specifically, we employ CLIPScore~\citep{hessel2021clipscore} with a $0.25$ cutoff which we selected by manually testing a sample of 100 QA pairs.

    \item \textbf{Spatial Relation Verification}
          We filter out any QA pairs that do not consist of spatial-reasoning questions similarly to the approach described in Subsection~\ref{subsec:dataset_analysis}.
          The difference here is that we classify based on the QA pair instead of the description.

\end{enumerate}

\par{}QA pairs were progressively filtered based on the aforementioned automated criteria, which are listed and applied in order of increasing computational requirements.
We also filter out QA pairs tied to images that we were unable to download.
This led us to drop around 50k samples mostly from PixMo-Cap.

\subsection{Human Evaluation}  
To further verify the quality of our dataset, we sample a representative subset of 400 samples from the dataset.
We describe the process by which we arrive at this number in Section \ref{appendix:human_evals_sample_size} in the appendix.  
We observe an error rate of around 4\% in the QA pairs, which we find reasonable for a synthetic dataset.

\subsection{Hallucination Mitigation}
In our initial studies, we identified the primary cause of hallucinations to be descriptions that did not contain any actual spatial relations.
This insight led us to implement an aggressive filtering strategy for such descriptions.
Following this refinement, hallucinations became manageable, with relation and object hallucination rates reduced to approximately 4\% and 3\%, respectively.

\subsection{Addressing Data Scarcity}

Our pipeline generates 455k samples with 3.4 million QA pairs, helping to address the lack of spatial reasoning data in VLM datasets. 
By offering diverse examples across various spatial relations and contexts, we improve VLMs' ability to learn and generalize spatial reasoning, leading to better performance on related tasks.

% \Freda{
%     1. We need a subsection here to describe the training objective---even if it's just a standard fine-tuning, it's worth stating things like we optimize the cross entropy loss on the predicted tokens with one equation or more.

%     2. This is probably already on your agenda, but Figure 1 could be made clearer by including more detailed examples.
% }

% \Mike{
%     1. Added the training obj section

%     2. I'm using the Figure in the introduction instead now.
% }

\subsection{Training Objective}

We train our VLMs by optimizing the cross-entropy loss between the model's predicted and the ground-truth text token probabilities, computing no loss on visual tokens.
Specifically, given an input image $I$ and its corresponding text input $X$ (i.e., a question), along with the target output sequence (i.e., an answer), $Y = \{y_1, y_2, \dots, y_T\}$, the model aims to minimize the negative log-likelihood of the target tokens given the inputs.

The training objective is defined as:

\begin{equation}
    \mathcal{L} = - \sum_{t=1}^{T} \log p_{\theta}(y_t \mid I, X, y_{<t})
    \label{eq:cross_entropy_loss}
\end{equation}

where $\theta$ represents the model parameters, ~$p_{\theta}(y_t \mid I, X, y_{<t})$ is the probability of generating the token $y_t$ at position $t$ given the image $I$, question $X$, and previous tokens $y_{<t}$.

\par{}We outline more training details in Subsection~\ref{subsec:training_procedure}.

\section{Experiments and Results}
\label{sec:experiments_results}

% Reduce space between columns
\setlength{\tabcolsep}{4pt}

\begin{table*}[htbp]
    \centering
    \small
    \begin{tabular}{lccccccccccccccc}
        \toprule
                      & \multicolumn{6}{c}{Spatial Reasoning Benchmarks} & \multicolumn{9}{c}{General Benchmarks} \\
        \cmidrule(lr){2-7} \cmidrule(lr){8-16}
        Model & \rot{VSR} & \rot{What's Up A} & \rot{What's Up B} & \rot{3DSRBench} & \rot{RealWorldQA} & \rot{Average}         
            & \rot{MMMU} & \rot{MMBench} & \rot{\shortstack{Hallusion\\Bench}} & \rot{TextVQA} & \rot{\shortstack{MME\\\textsubscript{perception}}} & \rot{\shortstack{MME\\\textsubscript{reasoning}}} & \rot{\shortstack{MME\\\textsubscript{code}}} & \rot{\shortstack{MME\\\textsubscript{commonsense}}} &
\rot{\shortstack{MME\\\textsubscript{numerical}}} \\
        \midrule
        Random & 50.0 & 25.0 & 25.0 & 20.9 & -- & 30.2 
            & -- & -- & -- & -- & -- & -- & -- & -- & -- \\
        Human Estimate & 95.4 & 100.0 & 100.0 & 95.7 & -- & 99.8
            & -- & -- & -- & -- & -- & -- & -- & -- & -- \\
        \midrule
        SpaRE-2B (Ours) & \textbf{80.8} & \textbf{93.4} & \textbf{95.1} & \textbf{54.4} & \textbf{63.5} & \textbf{77.6}         
            & \textbf{40.0} & 71.6 & 58.2 & \textbf{79.2} & 1467.9 & 432.4 & 108.5 & \textbf{110.1} & 39.5 \\
        Qwen2VL-2B & 70.3 & 44.6 & 79.1 & 46.5 & 58.6 & 59.8 
            & 34.0 & \textbf{72.0} & \textbf{61.2} & \textbf{75.0} & \textbf{1490.7} & \textbf{441.8} & \textbf{112.5} & 109.3 & 42.5 \\
        InternVL2-2B & 68.7 & 86.8 & 84.7 & 46.7 & 57.4 & 68.9 
            & 34.0 & 71.4 & 59.3 & 73.5 & 1442.2 & 423.6 & 92.5 & 108.6 & \textbf{45.0} \\
        \midrule
        SpaRE-7B (Ours) & \textbf{85.4} & \textbf{100.0} & \textbf{100.0} & \textbf{57.5} & \textbf{68.8} & \textbf{82.3} 
            & \textbf{51.0} & 78.6 & 56.3 & 80.5 & 1661.4 & \textbf{642.3} & 145.5 & \textbf{156.3} & 127.5 \\ 
        Qwen2VL-7B & 82.3 & 99.5 & 99.3 & 49.2 & 67.7 & 79.2
            & \textbf{51.0} & 79.9 & 59.9 & \textbf{81.7} & \textbf{1667.3} & 640.0 & \textbf{152.5} & 155.0 & \textbf{132.5} \\
        InternVL2-8B & 73.1 & 79.2 & 94.4 & 53.3 & 64.4 & 72.5 
            & 47.4 & \textbf{80.9} & \textbf{64.7} & 77.6 & 1649.6 & 572.1 & \textbf{152.5} & 147.1 & 87.5 \\
        LLaVA-NeXT-8B & 71.9 & 93.6 & 95.6 & 51.1 & 58.2 & 74.1
            & 46.7 & 72.5 & 39.0 & 65.6 & 1540.2 & 308.6 & 52.5 & 118.6 & 47.5 \\
        SpaceLLaVa\footnote{\url{https://hf.co/remyxai/SpaceLLaVa}} & 65.9 & 75.5 & 75.6 & 47.2 & 48.4 & 62.5
            & 35.3 & 66.5 & 43.9 & 32.4 & 1411.8 & 295.0 & 47.5 & 125.0 & 72.5 \\
        \midrule
        GPT-4o-mini & 74.0 & 75.0 & 90.0 & 39.1 & 56.0 & 66.4
            & -- & -- & -- & -- & -- & -- & -- & -- & -- \\
        GPT-4o & \textbf{79.0} & \textbf{100.0} & \textbf{100.0} & \textbf{45.3} &\textbf{61.0} & \textbf{77.9} 
            & -- & -- & -- & -- & -- & -- & -- & -- & -- \\
        \bottomrule
    \end{tabular}
    \caption{The performance of original vs SpaRE VLMs, along with competitor models across a wide selection of datasets divided into spatial reasoning and general benchmarks. The \textbf{best} score is emboldened.}
    \label{tab:main_results}
\end{table*}

% Reset column separation to default
\setlength{\tabcolsep}{6pt}

In this section, we detail the experimental setup used to evaluate our method as well as present and discuss the results.
Our experiments aim to evaluate the effectiveness of our approach and compare them to relevant baselines in enhancing VLMs' spatial understanding while maintaining general VL capabilities.

\subsection{Experimental Setup}

\subsubsection{Base VLM Selection}

We select \texttt{Qwen2-VL-2B-Instruct} and \texttt{Qwen2-VL-7B-Instruct} as our base VLMs for fine-tuning since they are leading open-source models in their respective size classes and thus provide strong foundations.
We do not experiment with larger models due to computing resource constraints.

% \subsubsection{Training Data Composition}

% In fine-tuning the base VLMs, we combine our synthetic dataset with VSR~\citep{liu-etal-2023-visual}, an existing dataset.
% We select a balanced training ratio across all the constituent datasets.

\subsubsection{Training Procedure}
\label{subsec:training_procedure}

To select our hyperparameters, we conducted a search.
We show our search and the selected hyperparameters for our \texttt{2B} and \texttt{7B} variants in Tables~\ref{tab:hyperparameter_tuning} and \ref{tab:hyperparameter_tuning_7b} respectively in the appendix.
We train with \texttt{bfloat16} precision for improved efficiency and a linear learning rate warm-up over the first $1{,}000$ steps, followed by a cosine decay schedule.
To stabilize training, we clipped gradients with a maximum norm of $1.0$.
We train the models with $5$ random seeds and report the mean results.

The training was performed on 4 NVIDIA A40 GPUs with $48$GB RAM.
For the \texttt{2B} model, all weights were trained, while for the \texttt{7B} model, we used LoRA~\citep{hu2021lora} to save memory.

% \freda{If we need space, the hyperparameter setups can be the first batch for moving to the appendix.}
% \mike{I've condensed it a bit by removing the list and using sentences}

\subsection{Evaluation Benchmarks}

To evaluate the effectiveness of our approach, we assessed the fine-tuned model on a range of benchmarks covering both spatial reasoning and general VL tasks.

\subsubsection{Spatial Reasoning Benchmarks}

We evaluate on these spatial reasoning benchmarks:

\begin{enumerate}[leftmargin=*] % [topsep=0pt]
    
    \item \textbf{Visual Spatial Reasoning (VSR)}~\citep{liu-etal-2023-visual}: Tests models' ability to a wide swath of understand 66 spatial relations in images through binary classification tasks.
    % \vspace{-8pt}

    \item \textbf{What's Up?}~\citep{kamath-etal-2023-whats}: Focuses on evaluating models' understanding of basic spatial relations, such as \emph{left}, \emph{right}, \emph{above}, \emph{below}, \emph{in-front}, and \emph{behind}.
    % \vspace{-8pt}

    \item \textbf{3D Spatial Reasoning Benchmark (3DSRBench)}~\citep{ma20243dsrbench}: Assesses models' capabilities in understanding 3D spatial relations in complex scenes.
    % \vspace{-8pt}

    \item \textbf{RealWorldQA}\footnote{\url{https://x.ai/blog/grok-1.5v}}: A dataset consisting of real-world images and questions requiring spatial reasoning to answer accurately.
    
\end{enumerate}

These benchmarks provide a comprehensive evaluation of spatial reasoning abilities across different types of spatial relationships and contexts.

\subsubsection{General VL Benchmarks}

To ensure spatial reasoning improvements do not sacrifice general performance, we evaluate on general VL benchmarks: MMMU~\citep{yue2024mmmu} for domain-specific multi-modal reasoning, MMBench~\citep{liu2024mmbench} for fine-grained vision–language skills, HallusionBench~\citep{Guan_2024_CVPR} for hallucinations, TextVQA~\citep{Singh_2019_CVPR_textvqa} for text-in-image reasoning, and MME~\citep{yin2023mme} for integrated multi-modal cognition.

These benchmarks assess the SpaRE models' general applicability and robustness relative to competing VLMs after spatial reasoning fine-tuning.

\subsection{Evaluation Metrics}
\iffalse
\Freda{I assume you're already working on these, but here are a few questions: 

1. You probably want a sentence describing what ``PT'' is (from our discussion: let's get a method name and make it less confusing)

2. Let's add a description of what Elo is as people may not be familiar with it. 

3. For closed-sourced models, we can add a few results from existing paper, and not necessarily to be all. 
}
\Mike{
I've addressed all these
}
\fi
We use accuracy as the primary metric for all spatial reasoning benchmarks. 
For multiple-choice tasks like What's Up, 3DSRBench, and RealWorldQA, we prompt the VLM to predict the correct option, and then apply string matching to compare the output with the ground truth. 
For binary classification tasks like VSR, we evaluate binary accuracy by predicting \textit{True} or \textit{False}.

Evaluations were conducted using VLMEvalKit~\citep{duan2024vlmevalkit} and our own code for unsupported benchmarks (VSR and What's Up). 
Some results are sourced from other works, as we detail in Section~\ref{appendix:results_sources} in the appendix. 
While most benchmarks use accuracy, MME uses a different scoring system. 
For MME, perception is scored out of 2,000, reasoning out of 800, and code, commonsense, and numerical tasks are scored out of 200 each.

% To maintain consistency, no dataset-specific fine-tuning or adjustments were applied beyond prompt modifications for different tasks.

\subsection{Baselines and Compared VLMs}
\label{sec:compared_vlms}

We compare our SpaRE models to multiple VLMs.  
For baselines, we include a random baseline for reference, which assigns answers uniformly at random, and a human estimate baseline from benchmark authors where available.
We evaluate leading open-source models, including Qwen2VL~\citep{wang2024qwen2vl}, InternVL2~\citep{chen2024farinternvl1_5}, LLaVA-NeXT~\citep{liu2024visual}, and SpaceLLaVa~\cite{chen2024spatialvlm}, a model specifically optimized for quantitative spatial reasoning. 
We also compare against proprietary models: GPT-4o and GPT-4o-mini~\cite{achiam2023gpt}.

\subsection{Results and Discussion}
\label{sec:results_discussion}

\subsubsection{Spatial Reasoning Performance}

Our fine-tuned models exhibit substantial improvements across the spatial reasoning benchmarks. 
Specifically, the average accuracy of the \texttt{2B} and \texttt{7B} variants increase by around \textbf{9\%} and \textbf{3\%} across these tasks. 
These gains demonstrate the effectiveness of our synthetic spatial reasoning data in enhancing the model's spatial reasoning abilities. 
By incorporating explicit spatial relationships and diverse spatial contexts into the training data, the fine-tuned model developed a more robust understanding of spatial concepts. 
This suggests that the scarcity of spatial reasoning data in existing datasets was a key factor limiting the spatial capabilities of VLMs.

\subsubsection{General VL Performance}

The results show that the fine-tuned model performs on par with the original models in general VL tasks, with minimal differences. 
This suggests that incorporating synthetic spatial reasoning data does not harm overall capabilities. 
During QA generation, we observe \emph{benign} hallucinations—QA pairs relevant to the image but unrelated to spatial reasoning. 
Including these in training helps prevent overfitting and preserves general performance. 
The ability to enhance spatial reasoning while maintaining broad VL competence highlights the effectiveness of our data augmentation approach.
% The balanced mixing ratio of the synthetic and original data likely contributed to this outcome by preventing overfitting to spatial reasoning tasks.

\subsubsection{Impact of Synthetic Spatial Reasoning Data}

The significant improvements in spatial reasoning tasks come from our effective use of synthetic data.  
By training on a wide range of spatial relations and situations, the model learns to understand and reason about space more accurately.  
We create QA pairs from detailed image captions, covering many types of spatial relationships.  
This variety helps the model apply spatial reasoning to new situations, even those not seen in training.  
However, the quality of the synthetic data is crucial for training a strong model.  
By carefully designing prompts and using a powerful, but fast LLM, we generate high-quality QA pairs that correctly reflect spatial relationships in images.  
Still, any shortcomings or biases in the LLM could affect the quality of the synthetic data.

\subsubsection{Generalization to Real-World Scenarios}

\texttt{SpaRE} models improve by 4.9\% and 1.1\% on the original \texttt{2B} and \texttt{7B} models in RealWorldQA, showing that its spatial reasoning skills extend to real-world data, not just synthetic or controlled settings.
The improvements on the RealWorldQA benchmark show that the model can apply spatial reasoning to real-world images and questions.  
RealWorldQA includes everyday scenes and tests the application of spatial reasoning to solve those tasks.
This shows promise for our approach beyond academic settings. 
This is especially useful for applications like robotics, where models must understand complex spatial relationships in unpredictable environments.

\subsection{Qualitative Analysis}

\begin{figure}[ht]
    \centering

    % Subfigure for the image
    \begin{subfigure}[t]{0.8\linewidth}
        \centering
        \includegraphics[width=\linewidth]{images/man_in_living_room.jpg}
        % \caption{Image used for the question-answer comparison.}
        \label{fig:eval_example_image}
    \end{subfigure}

    % Subfigure for the table
    \begin{subfigure}[t]{\linewidth}
        \centering
        \small
        \begin{tabular}{m{0.325\linewidth}p{0.3\linewidth}p{0.075\linewidth}p{0.05\linewidth}}
            \toprule
            {Question}                                      & {Models}  & {Answers}                                       & \textbf{}                      \\
            \midrule
            \multirow{5}{=}{If I stand at the person's position facing where it is facing, is the table on the left or right of me?} & {SpaRE-7B} & Right & \textbf{\color{green}{\cmark}} \\
            \cmidrule(lr){2-4}
                                                                   & Qwen2-VL-7B    & Left     & \textbf{\color{red}{\xmark}}   \\
            \cmidrule(lr){2-4}
                                                                   & Qwen2-VL-72B    & Left     & \textbf{\color{red}{\xmark}}   \\
            \cmidrule(lr){2-4}
                                                                   & InternVL2-76B & Left                    & \textbf{\color{red}{\xmark}}   \\
            \cmidrule(lr){2-4}
                                                                   & GPT-4o          & Left & \textbf{\color{red}{\xmark}}   \\
            \bottomrule
        \end{tabular}
        % \caption{Qualitative comparison of answers provided by different models for the given question. Our model provides the correct spatial relation answer.}
        \label{tab:qa_qualitative_comparison}
    \end{subfigure}

    \caption{
    Comparison of answers provided by different VLMs to a spatial reasoning question.
    }

    \label{fig:qualitative}
\end{figure}

In Figure~\ref{fig:qualitative}, when asked, \textit{``...is the table on the left or right of me?''}, \texttt{SpaRE-7B} correctly answers \textit{``Right''}, while the compared VLMs provide incorrect responses. 
This question is difficult because while the table is on the left, in the viewer's and thus VLM's perspective, it is to the right from the perspective of the person shown.
This example demonstrates the practical improvements achieved with our method. 
While questions like this remain challenging for SpaRE VLMs, as discussed shortly, they significantly outperform existing alternatives.
We show more examples in Figure~\ref{fig:model_usage_examples} in the appendix.

\subsection{Error Analysis}
While our performance on VSR and What's Up is strong, we see relatively weaker performance on 3DSRBench. 
We observe, consistent with \citet{zhang2024visionlanguagemodelsrepresentspace}, that VLMs struggle with \emph{empathetic} spatial reasoning.
That means that they fail to adopt the perspectives of others when reasoning about spatial relations. 
This egocentric bias, originating from source datasets, is also present in our synthetic dataset. 
Addressing this will likely require datasets that capture manually annotated information about different frames of reference in a scene.

\section{Conclusion}
\label{sec:conclusion}

In this work, we tackled the lack of spatial reasoning data in VL datasets by generating synthetic QA pairs from hyper-detailed image captions using LLMs.  
Our approach greatly improves the spatial reasoning of VLMs, as shown by strong gains across benchmarks, without hurting general VL performance.  
By using rich spatial descriptions to create diverse and accurate QA pairs, we provided the data needed for VLMs to learn and apply spatial reasoning effectively.  
Our results emphasize the importance of data quality and diversity in training robust VLMs and open new directions for research in robotics, navigation, and extended reality.  
We hope this work encourages further exploration of synthetic data to address VLM limitations, helping build more capable and versatile multi-modal AI systems.

\section{Limitations}
\label{sec:limitations_and_future_work}

% Speak about the limitations of using synthetic data

\begin{figure}[h]
    \centering
    \includegraphics[width=0.75\linewidth]{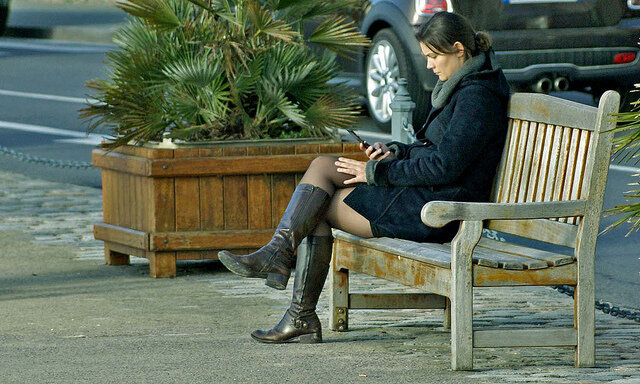}
    \caption{Ambiguity of spatial relations without an explicit frame of reference: is the plant to the \emph{right} or \emph{left} of the bench from the \emph{viewer's} or \emph{woman's} perspective?}
    \label{fig:ambiguity}
\end{figure}

% \freda{To make it safe, I suggest naming this section just by ``Limitations.'' We may move some future work discussion to the conclusion section if space allows.} 
While our approach significantly improves the spatial reasoning abilities of VLMs, we recognize some limitations.
For instance, spatial descriptions can be ambiguous without clear frames of reference (FOR) \citep{Levinson_2003, liu-etal-2023-visual}.
Moreover, we and \citet{zhang2024visionlanguagemodelsrepresentspace} observe that current VLMs fall short in capturing different FORs during spatial reasoning.
As shown in Figure~\ref{fig:ambiguity}, whether the plant is to the \emph{right} or \emph{left} of the bench depends on whether a viewer-centric or object-centric perspective is used—without specifying the FOR, it is hard to determine the intended meaning.

In future work, we plan to address this challenge by incorporating explicit FOR annotations into our synthetic data generation pipeline.
We also intend to explore more efficient methods for generating synthetic data with large language models and to adapt our approach to other languages with richer morphology.

\section{Ethics Statement}
\label{sec:ethics}

Our work uses synthetic data generated from hyper-detailed image descriptions to improve spatial reasoning in VLMs. 
The source data is publicly available and does not contain sensitive information. 
However, since improved spatial reasoning can affect applications like robotics or navigation, it is important to test these models thoroughly before deployment. 
We also note that the language models used to generate synthetic data may exhibit biases from their training data. 
To help the community verify and build on this work, we plan to release our code and data under open-access terms.

% \section*{Acknowledgments}
% Resources used in preparing this research were provided, in part, by the Province of Ontario, the Government of Canada through CIFAR, and companies sponsoring the Vector Institute \url{www.vectorinstitute.ai/#partners}.

\bibliography{custom.bib}

\newpage

\appendix

\section{Experiments}
\label{appendix:experiments}

\subsection{Hyperparameter Tuning}

To achieve the reported batch sizes, we employed gradient accumulation. 
Tables~\ref{tab:hyperparameter_tuning} and \ref{tab:hyperparameter_tuning_7b} summarize the hyperparameter search and the selected values for training the \texttt{SpaRE-2B} and \texttt{SpaRE-7B} models respectively. 
The search includes variations in learning rate (LR), batch size, epochs, optimizer, warm-up steps, dropout rate, and learning rate schedule. 
For both models, the optimal values were chosen based on performance on a held-out subset.

\begin{table}[ht]
    \small
    \centering
    \begin{tabular}{lll}
        \toprule
        {Hyperparameter} & {Tried}                         & {Selected}  \\
        \midrule
        Learning rate      & $1 \times 10^{-5}$, $3 \times 10^{-5}$ & $3 \times 10^{-5}$ \\
        Batch size   & 32, 64                            & 64                 \\
        Epochs        & 1, 2, 3                              & 2                  \\
        Optimizer               & AdamW, AdamW8bit                    & AdamW8bit              \\
        Warm-up steps           & 500, 1000, 1500                        & 1000               \\
        Dropout            & 0, 0.1                                 & 0.1                \\
        Schedule            & Linear, Cos, Fixed                                 & Cos                \\
        \bottomrule
    \end{tabular}
    \caption{Summary of hyperparameter search and the values selected for \texttt{SpaRE-2B} training.}
    \label{tab:hyperparameter_tuning}
\end{table}

\begin{table}[ht]
    \small
    \centering
    \begin{tabular}{lll}
        \toprule
        {Hyperparameter} & {Tried}                         & {Selected}  \\
        \midrule
        Learning rate      & $3 \times 10^{-4}$, $1 \times 10^{-4}$ & $1 \times 10^{-4}$ \\
        Batch size   & 32, 64                            & 64                 \\
        Epochs        & 1, 2, 3                              & 2                  \\
        Optimizer               & AdamW, AdamW8bit                    & AdamW              \\
        Warm-up steps           & 500, 1000, 1500                        & 1000               \\
        Dropout            & 0, 0.1                                 & 0.1                \\
        Schedule            & Linear, Cos, Fixed                                 & Cos                \\
        \bottomrule
    \end{tabular}
    \caption{Summary of hyperparameter search and the values selected for \texttt{SpaRE-7B} training.}
    \label{tab:hyperparameter_tuning_7b}
\end{table}

\section{Pipeline Ablations}
\label{appendix:ablation_studies}

We subsequently describe the ablations that we carried out for QA-pair generation and source dataset filtering.

\subsection{QA Generation Ablations}

We perform ablations on $100$ samples using the following \texttt{Qwen2.5} instruct models to generate synthetic datasets from source descriptions: \texttt{0.5B}, \texttt{1.5B}, \texttt{3B}, and \texttt{7B}.
We manually check the QA pairs in each sample.

\begin{table}[H]
    \centering
    \begin{tabular}{lcc}
        \toprule
        Model & Spatial relevance & QA pairs \\
        \midrule
        0.5B  & 0.17 & 4.8 \\
        1.5B  & 0.93 & 13.6 \\
        3B    & 0.89 & 17 \\
        7B    & 0.86 & 14 \\
        \bottomrule
    \end{tabular}
    \caption{QA pair generation ablation study results for \texttt{Qwen2.5} instruct models.}
    \label{tab:qa_generation_model_ablation_results}
\end{table}

The \texttt{0.5B} model performs poorly and is discarded. 
Among the remaining models, the key difference is the number of QA pairs generated. 
The \texttt{3B} model generates the most QA pairs, making it the preferred choice, as we can filter non-spatial questions later. 
While the spatial relevance rate isn't the highest, the questions generated are still relevant, and including these samples in the dataset helps prevent overfitting to spatial reasoning.
% Also, compared to \texttt{7B}, it runs about X times faster.
We stop at \texttt{7B} due to computational resource limitations, as the smaller model sizes already perform decently.

\subsection{Dataset Filtering Ablations}

We run ablations on the following \texttt{Qwen2.5} instruct models for classifying source descriptions with spatial information present: \texttt{0.5B}, \texttt{1.5B}, \texttt{3B}, \texttt{7B}.
We manually check $100$ classified samples.

\begin{table}[H]
    \centering
    \small
    \begin{tabular}{lccccc}
        \toprule
        Model & Compute Time & Acc & P & R & F1 \\
        \midrule
        0.5B  & 1x           & 0.33  & 1 & 0.20 & 0.33 \\
        1.5B  & 1.26x        & 0.60  & 1 & 0.56 & 0.71 \\
        3B    & 1.42x        & 0.57  & 1 & 0.54 & 0.70 \\
        7B    & 1.49x        & 0.63  & 1 & 0.58 & 0.73 \\
        \bottomrule
    \end{tabular}
    \caption{
    Dataset-filtering ablation study results for \texttt{Qwen2.5} instruct models. 
    Acc refers to accuracy, P refers to precision, R refers to recall, and F1 refers to F1-score.}
    \label{tab:data_filtering_model_ablation_results}
\end{table}

We stop at \texttt{7B} due to computational resource limitations, as the smaller model sizes already perform well enough for our requirements.
Additionally, we note that precision is the most important metric, as passing in sample descriptions without actual spatial relations—due to false positives—leads to severe hallucinations, negatively impacting downstream performance.

\section{Qualitative Analysis}
\label{appendix:qualitative_analysis}

\subsection{Response Generation}
\begin{figure*}[ht]
\centering

% First Example
\begin{subtable}{\textwidth}
    \centering
    \begin{tabular}{c}
        \includegraphics[width=0.4\linewidth]{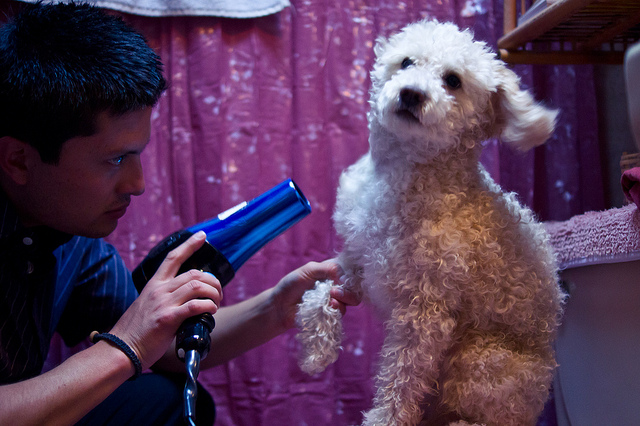} \\
    \end{tabular}
    \vspace{2mm}
    
    \begin{tabular}{c}
        \textbf{Question:} The hair drier is facing away from the person. True or False? \\
    \end{tabular}
    \vspace{2mm}
    
    \begin{tabular}{ccc}
        \toprule
        Model & Prediction & Correct? \\
        \midrule
        SpaRE-2B & True & \color{green}{\cmark} \\
        Qwen2VL-2B & False & \color{red}{\xmark} \\
        InternVL2-2B & No & \color{red}{\xmark} \\
        GPT-4o-mini & False & \color{red}{\xmark} \\
        \bottomrule
    \end{tabular}
    \caption{Example 1 from VSR}
\end{subtable}

\vspace{5mm}

% Second Example
% \begin{subtable}{\textwidth}
%     \centering
%     \begin{tabular}{c}
%         \includegraphics[width=0.4\linewidth]{images/cars_and_dog_on_sidewalk.jpg} \\
%     \end{tabular}
%     \vspace{2mm}
    
%     \begin{tabular}{c}
%         \textbf{Question:} If I stand at the lime green car's position facing where it is facing, \\is the dog in front of me or behind me? \\
%     \end{tabular}
%     \vspace{2mm}
    
%     \begin{tabular}{ccc}
%         \toprule
%         Model & Answer & Correct? \\
%         \midrule
%         SpaRE-2B & In front of me & \color{green}{\cmark} \\
%         Qwen2VL-2B & Behind me & \color{red}{\xmark} \\
%         InternVL2-2B & Behind me & \color{red}{\xmark} \\
%         \bottomrule
%     \end{tabular}
%     \caption{Example 2 from 3DSRBench}
% \end{subtable}

% Third Example
\begin{subtable}{\textwidth}
    \centering
    \begin{tabular}{c}
        \includegraphics[width=0.4\linewidth]{images/person_facing_laptop.jpg} \\
    \end{tabular}
    \vspace{2mm}
    
    \begin{tabular}{c}
        \textbf{Question:} Which object is the person facing towards, the laptop or the TV? \\
    \end{tabular}
    \vspace{2mm}
    
    \begin{tabular}{ccc}
        \toprule
        Model & Answer & Correct? \\
        \midrule
        SpaRE-2B & Laptop & \color{green}{\cmark} \\
        Qwen2VL-2B & TV & \color{red}{\xmark} \\
        InternVL2-2B & TV & \color{red}{\xmark} \\
        GPT-4o-mini & TV & \color{red}{\xmark} \\
        \bottomrule
    \end{tabular}
    \caption{Example 2 from 3DSRBench}
\end{subtable}

\caption{For our qualitative analysis, each sub-figure contains an image, a corresponding question, different models’ responses, and their correctness ({\color{green}{\cmark}} or {\color{red}{\xmark}}).}
\label{fig:model_usage_examples}
\end{figure*}

We show more examples of SpaRE models compared to similar VLMs in Figure \ref{fig:model_usage_examples}.

\section{Prompts}
\label{appendix:prompts}

We show the prompts we used for the VQA dataset analysis and QA pair generation. 

\subsection{Dataset Analysis Prompt}

In Table~\ref{tab:spatial_relation_check_prompt}, we present the prompt used for analyzing the VQA dataset to identify spatial relations within descriptions. The prompt asks the model to determine whether a given description contains a spatial relation, helping to filter relevant samples for further analysis or QA pair generation.

\begin{table}[h]
    \centering
    \small
    \begin{tabular}{p{0.9\linewidth}}
        \toprule
        Determine if the description provided below contains a spatial relation: ~\textbf{\{description\}} \\
        \bottomrule
    \end{tabular}
    \caption{Prompt for identifying spatial relations in descriptions.}
    \label{tab:spatial_relation_check_prompt}
\end{table}

\subsection{QA Generation Prompt}

In Table~\ref{tab:spatial_relation_qa_generation_prompt}, we present the prompt used for generating question-answer (QA) pairs from image descriptions focused on spatial relations. 
The prompt instructs the model to generate a JSON list of QA pairs, with each question centered on spatial details such as object positions, orientations, distances, and interactions. 
The output format is specified, and the prompt guides the model to focus solely on spatial relationships while excluding non-relevant questions.

\begin{table}[H]
    \centering
    \small
    \begin{tabular}{p{0.9\linewidth}}
        \toprule
        Generate a JSON list of question-answer pairs based on the detailed image description below. The questions should exclusively focus on spatial relations between objects, including their positions, orientations, distances, and any relevant interactions that define their relative locations. Avoid questions outside of spatial details. \\ \\

        The output should look like:                                                                                                                                                                                                                                                                                                                  \\

        [                                                                                                                                                                                                                                                                                                                                             \\
        \hspace{1em} \{"question": <question>, "answer": <answer>\},                                                                                                                                                                                                                                                                                  \\
        \hspace{1em} \dots                                                                                                                                                                                                                                                                                                                            \\
        ]                                                                                                                                                                                                                                                                                                                                             \\ \\

        For spatial relations, consider asking about:                                                                                                                                                                                                                                                                                                 \\
        - {Positions and Directions}: Where objects are located (e.g., left, right, above, below).                                                                                                                                                                                                                                             \\
        - {Relative Distances and Proximity}: How close or far objects are from each other.                                                                                                                                                                                                                                                    \\
        - {Orientations and Angles}: Any notable angles or orientations of objects (e.g., tilted, rotated).                                                                                                                                                                                                                                    \\
        - {Foreground and Background Layers}: Which elements are in the foreground, middle ground, or background.                                                                                                                                                                                                                              \\
        - {Boundaries and Edges}: How objects align with edges or blend into the background.                                                                                                                                                                                                                                                   \\
        - {Interaction of Shadows and Reflections}: Shadow placement relative to objects, or how objects reflect on surfaces.                                                                                                                                                                                                                  \\
        - {Overlapping and Layering}: If objects overlap or are layered, which ones appear on top or behind.                                                                                                                                                                                                                                   \\
        - {Scale and Size Comparisons}: Relative sizes between objects based on spatial cues.                                                                                                                                                                                                                                                  \\ \\

        Output only the JSON, starting with `[' and ending with `]'.                                                                                                                                                                                                                                                                                  \\ \\

        Image description: \textbf{\{description\}}                                                                                                                                                                                                                                                                                                   \\
        \bottomrule
    \end{tabular}
    \caption{Prompt to generate QA pairs from a description focused on spatial relations.}
    \label{tab:spatial_relation_qa_generation_prompt}
\end{table}

\section{Human Evaluation}  
\label{appendix:human_evals}

A research team member manually validated a sample of 400 entries from the dataset to assess quality and accuracy. 
The selection process is detailed below.

\subsection{Sample Size}
\label{appendix:human_evals_sample_size}

We determined the sample size to use with the finite population formula:  

\[
n = \frac{N Z^2 p(1-p)}{E^2 (N-1) + Z^2 p(1-p)}
\]
\\

where \( N = 455,494 \) (dataset size), \( Z = 1.96 \) (95\% confidence level), \( p = 0.5 \) (maximum variance), and \( E = 0.05 \) (margin of error). Substituting these values, we obtain \( n \approx 384 \), which we round up to 400 for robustness.

\section{Sources of Results}
\label{appendix:results_sources}

For the main results table (Table~\ref{tab:main_results}), values are computed using VLMEvalKit~\citep{duan2024vlmevalkit} and our custom evaluation code for unsupported benchmarks. 
Below, we outline cases where external sources were used or special handling was required.

Random baseline numbers for VSR~\citep{liu-etal-2023-visual}, What's Up~\citep{kamath-etal-2023-whats}, and 3DSRBench~\citep{ma20243dsrbench} were taken from the benchmark authors. 
For VSR and What’s Up, we used our evaluation code for all models.

For GPT-4o-mini and GPT-4o, we sampled 100 examples from VSR, What's Up A, and What's Up B to manage costs.
For 3DSRBench, we used the results reported in~\citet{ma20243dsrbench} for these models.

We did not fill in results for GPT-4o and GPT-4o-mini under general VL benchmarks, as the purpose of those benchmarks was to confirm that spatial reasoning fine-tuning does not significantly degrade general VL task performance.

\section{Datasets}
\label{appendix:datasets}

\subsection{Hyper-Detailed Descriptions}

\begin{figure*}[h]
    \centering
    \begin{minipage}{\textwidth}
        \centering
        \includegraphics[width=3in]{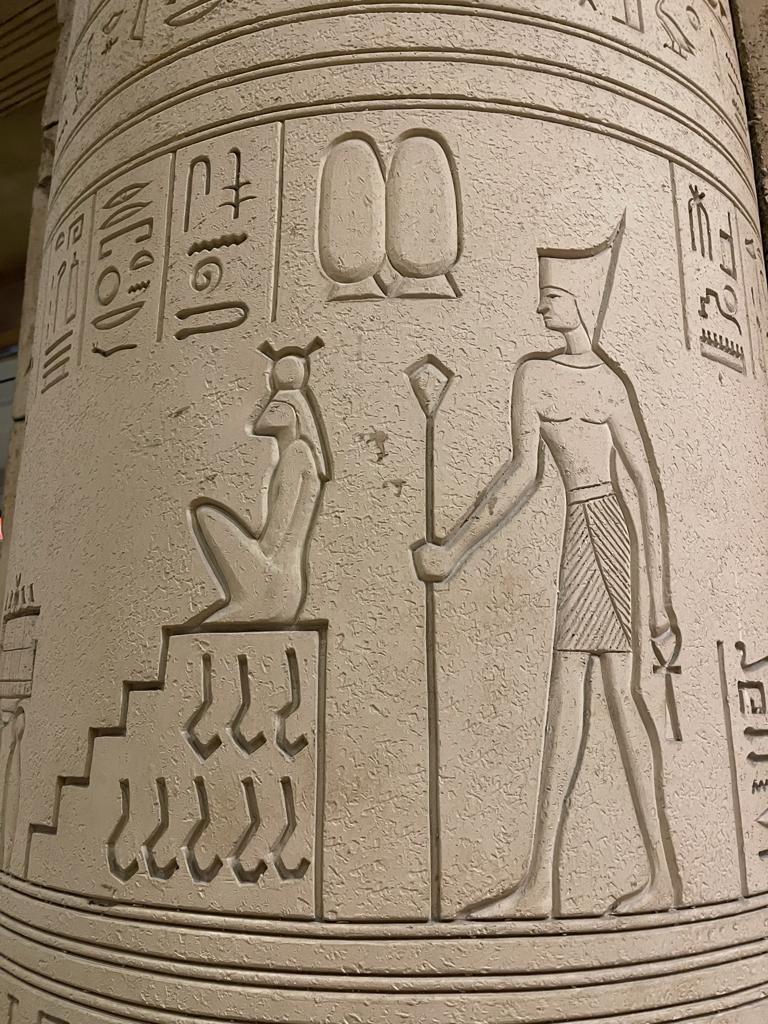}
        \vspace{6pt}
        \small
        \\
        A {low-angle close-up} view of an off-white pillar with Egyptian-style illustrations and hieroglyphics carved into it. The carved illustration in the middle of the image depicts a person holding a long staff with a diamond shaped object at the top of it \emph{in their right hand} as they are placing the bottom of the staff on the ground, and a cross-shaped object with a curved handle \emph{on top of} it \emph{in their left hand}. The person is facing \emph{the left side of the image} with their right foot ahead of their left. \emph{To the left of the person} is a being sitting \emph{on top of} a small set of stairs. The being is sitting with its knees bent and its feet \emph{in front of their body}, its feet and rear are both \emph{touching the surface}, and its hands are placed \emph{in its lap}. It is wearing a head dress that is long \emph{in the back} with a circular object placed \emph{on top}. The being doesn't look human nor is it sitting like a human. There are hieroglyphs and shapes carved in the pillar \emph{around and above} this illustration.
    \end{minipage}
    \caption{
    An example from DOCCI, one of the hyper-detailed image-captioning datasets that we extract QA pairs from. 
    We \emph{italicize} spatially relevant words for emphasis.}
    \label{hyper_detailed_captions}
\end{figure*}

\begin{figure*}[h]
    \centering
    \begin{minipage}{\textwidth}
        \centering
        \includegraphics[width=3in]{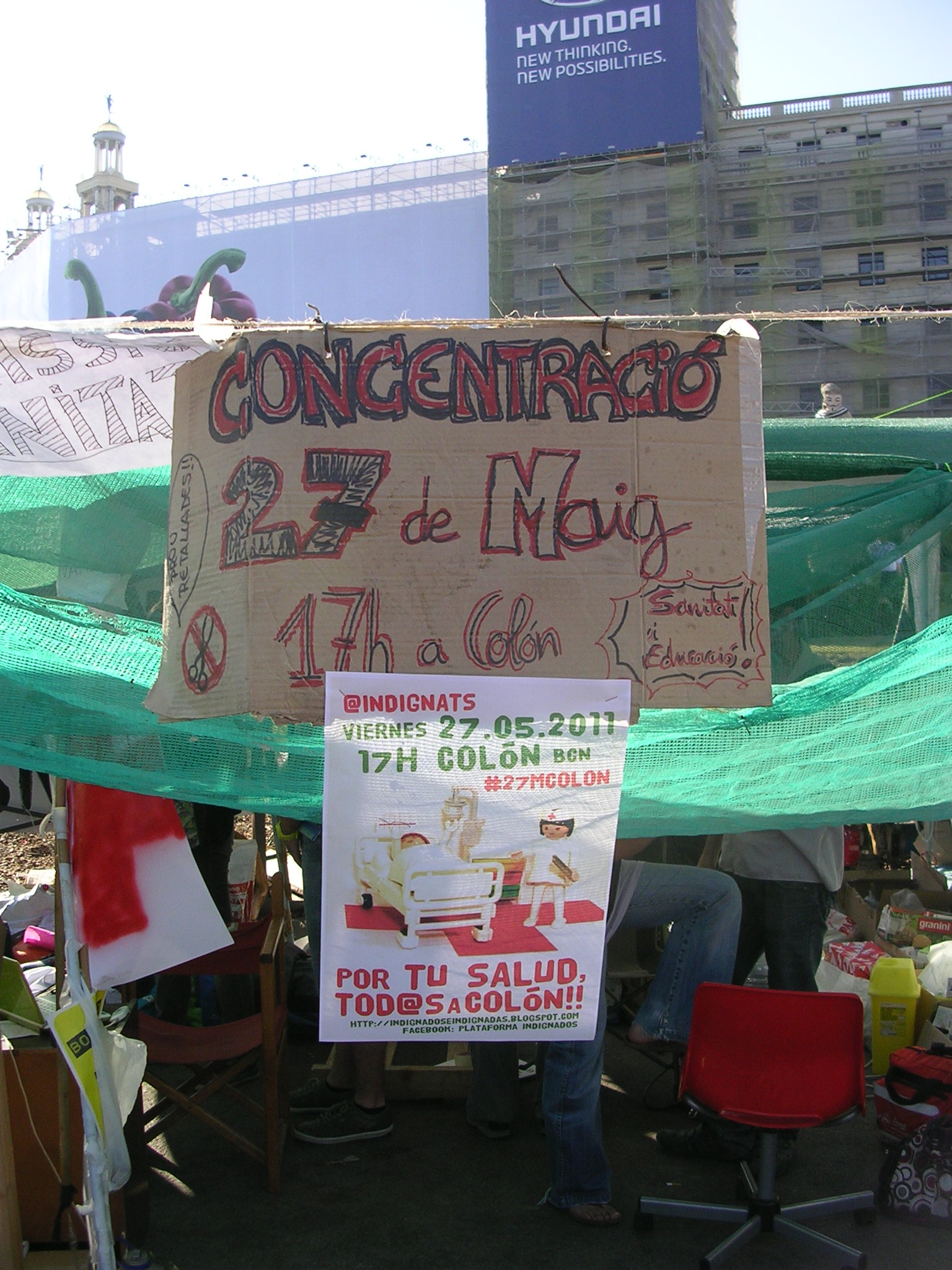}
        \vspace{6pt}
        \small
        \\
         The image is outside a building. There is a tent. There is a board and banner \emph{on} the tent. \emph{Under} the tent there are chairs, people, and some other stuff. In the \emph{background} there is a building. The sky is clear.
    \end{minipage}
    \caption{
    An example from Localized Narratives, one of the hyper-detailed image-captioning datasets that we extract QA pairs from. 
    We \emph{italicize} spatially relevant words for emphasis.}
    \label{hyper_detailed_captions_ln}
\end{figure*}

\begin{figure*}[h]
    \centering
    \begin{minipage}{\textwidth}
        \centering
        \includegraphics[width=3in]{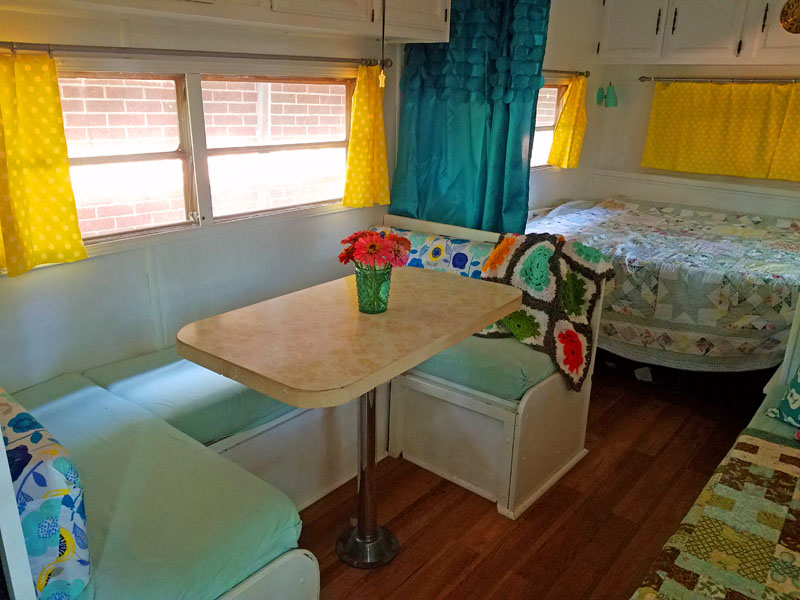}
        \vspace{6pt}
        \small
        \\
         The image captures the cozy interior of a camper van on a bright sunny day. Dominating the scene is a booth-like, U or C-shaped seating area upholstered in light teal or mint green cushions, accented by colorful throw pillows. This seating \emph{encircles} a tan rectangular table \emph{supported} by a chrome pole. \emph{Atop} the table rests a green vase filled with red flowers, each featuring prominent yellow centers. The camper is adorned with mustard yellow polka-dotted curtains framing the windows, which allow views of bricks outside, indicating the presence of a nearby building. \emph{Towards the back} of the camper, there is an area separated by teal curtains, which likely serves as a bedroom featuring a rounded bed draped in a quilt with light green and white pastels. The camper's interior is enhanced by white cupboards running \emph{along} the upper portions, providing ample storage. A crocheted throw with multicolored squares is casually draped over one of the bench seats, hinting at the occupant's knack for needlework. \emph{On} the dark brown wooden floor, the slightly cramped yet inviting space emphasizes both comfort and practical use of space \emph{in} this quaint, mobile home.
    \end{minipage}
    \caption{
    An example from PixMo-Cap, one of the hyper-detailed image-captioning datasets that we extract QA pairs from. 
    We \emph{italicize} spatially relevant words for emphasis.}
    \label{hyper_detailed_captions_pixmo}
\end{figure*}

We show examples of image-description pairs from DOCCI~\citep{OnoeDocci2024}, Localized Narratives~\citep{PontTuset_eccv2020}, and PixMo-Cap~\cite{deitke2024molmopixmoopenweights} in Figures \ref{hyper_detailed_captions}, \ref{hyper_detailed_captions_ln}, and \ref{hyper_detailed_captions_pixmo}.

\section{Basic Spatial Relation Taxonomy}
\label{appendix:taxonomy}

We define a basic taxonomy of spatial relations, inspired by \citet{liu-etal-2023-visual}, categorizing them into coarse- and fine-grained keywords. 
This framework helps analyze the distribution and use of spatial relations in VQA datasets.

Table~\ref{tab:basic_spatial_relations_taxonomy} outlines the taxonomy with key statistics. 
Each spatial relation (SR) is paired with relevant keywords (K), distinguishing broad categories from specific instances. 
The table also shows the percentage of each keyword within its SR group (K \%), the overall share of each SR in the dataset (SR \%), and its relative frequency across datasets.

The datasets analyzed are listed in Table~\ref{tab:dataset_analysis}. 
This taxonomy standardizes spatial relation interpretation, promoting a structured approach to spatial reasoning in VQA and related tasks.

\small
\onecolumn
\begin{longtable}{llrrH}
\caption{
Our taxonomy of spatial relations shows relations, sub-keywords, and their percentages and frequencies observed in selected VQA datasets. 
The covered datasets are detailed in Table~\ref{tab:dataset_analysis}. \\
\textbf{Spatial relation (SR)}: High-level relation category. \\
\textbf{Keyword (K)}: Sub-keyword representing the SR. \\
\textbf{K \%}: Percentage of the sub-keyword among all keywords in its SR group. \\
\textbf{SR \%}: Percentage of the high-level spatial relation in the dataset. \\
\textbf{Dataset freq.}: Relative frequency (in \%) of the spatial relation in the datasets. \\
}
\label{tab:basic_spatial_relations_taxonomy} \\
\toprule
Spatial relation & Keyword & K (\%) & SR (\%) & Dataset freq. (\%) \\
\midrule
\endfirsthead

\multicolumn{5}{c}{{\bfseries Table \thetable\ (continued)}} \\
\toprule
Spatial relation (SR) & Keyword (K) & K \% & SR \% & Dataset freq. (\%) \\
\midrule
\endhead

\midrule \multicolumn{5}{r}{{Continued on next page}} \\ 
\endfoot

\bottomrule
\endlastfoot

% ------------------------ Group: left (7 keywords) ------------------------
\multirow{7}{*}{left} 
  & left                & 7.41 & 19.28 & 6.63 \\
  & at the left         & 1.27 &       &      \\
  & on the left         & 1.24 &       &      \\
  & to the left of      & 3.04 &       &      \\
  & left of             & 3.75 &       &      \\
  & left side of        & 1.31 &       &      \\
  & at the left side of & 1.25 &       &      \\
\midrule

% ------------------------ Group: right (7 keywords) ------------------------
\multirow{7}{*}{right} 
  & right                & 8.04  & 21.65 & 7.44 \\ 
  & at the right        & 1.42  &       &      \\
  & on the right        & 1.75  &       &      \\
  & to the right of     & 3.40  &       &      \\
  & right of            & 4.03  &       &      \\
  & right side of       & 1.59  &       &      \\
  & at the right side of& 1.42  &       &      \\
\midrule

% ------------------------ Group: above (6 keywords) ------------------------
\multirow{7}{*}{above} 
  & above               & 1.64  & 2.35 & 0.81 \\ 
  & directly above      & \mytexttilde0.00  &  \\ 
  & over                & 0.37  &  \\ 
  & over the            & 0.34  &  \\ 
  & upward of           & \mytexttilde0.00  &  \\ 
  & overlying           & \mytexttilde0.00  &  \\ 
\midrule

% ------------------------ Group: on (10 keywords) ------------------------
\multirow{7}{*}{on} 
  & on                  & 9.29  & 13.04 & 4.48 \\ 
  & on top of           & 1.72  &  \\ 
  & atop                & \mytexttilde0.00  &  \\ 
  & on the top of       & 0.04  &  \\ 
  & on top              & 1.72  &  \\ 
  & lying on            & 0.04  &  \\ 
  & sitting on          & 0.22  &  \\ 
  & positioned on       & \mytexttilde0.00  &  \\ 
  & placed on           & \mytexttilde0.00  &  \\ 
  & overlaying on       & \mytexttilde0.00  &  \\ 
\midrule

% ------------------------ Group: below (10 keywords) ------------------------
\multirow{7}{*}{below} 
  & below               & 1.42  & 8.32 & 2.86 \\ 
  & under               & 1.98  &  \\ 
  & beneath             & 1.29  &  \\ 
  & directly below      & \mytexttilde0.00  &  \\ 
  & down                & 0.13  &  \\ 
  & underneath          & 0.04  &  \\ 
  & under the           & 1.98  &  \\ 
  & below the           & 1.40  &  \\ 
  & lower               & 0.06  &  \\ 
  & down from           & \mytexttilde0.00  &  \\ 
\midrule

% ------------------------ Group: front (9 keywords) ------------------------
\multirow{7}{*}{front} 
  & front               & 3.66  & 10.60 & 3.64 \\ 
  & in front of         & 3.40  &  \\ 
  & in the front of     & 0.02  &  \\ 
  & directly in front of & \mytexttilde0.00  &  \\ 
  & front of            & 3.51  &  \\ 
  % & looking at          & \mytexttilde0.00  &  \\ 
  & confronting         & \mytexttilde0.00  &  \\ 
\midrule

\newpage

% ------------------------ Group: back (8 keywords) ------------------------
\multirow{7}{*}{back} 
  & back                & 0.30  & 3.75 & 1.29 \\ 
  & behind              & 3.04  &  \\ 
  & at the back of      & 0.19  &  \\ 
  & in back of         & \mytexttilde0.00  &  \\ 
  & directly behind    & \mytexttilde0.00  &  \\ 
  & rear of            & \mytexttilde0.00  &  \\ 
  & backing onto       & \mytexttilde0.00  &  \\ 
  & back of            & 0.22  &  \\ 
\midrule

% ------------------------ Group: near\_close (7 keywords) ------------------------
\multirow{7}{*}{near\_close} 
  & near                & 0.75  & 1.12 & 0.39 \\ 
  & near to             & 0.02  &  \\ 
  & nearby              & 0.02  &  \\ 
  & close to            & 0.32  &  \\ 
  & close by            & \mytexttilde0.00  &  \\ 
  & in proximity to     & \mytexttilde0.00  &  \\ 
  & within sight of     & \mytexttilde0.00  &  \\ 
\midrule

% ------------------------ Group: far (6 keywords) ------------------------
\multirow{7}{*}{far} 
  & far                 & 1.08  & 2.07 & 0.71 \\ 
  & far from            & 0.28  &  \\ 
  & far away from       & 0.71  &  \\ 
  & farther than        & \mytexttilde0.00  &  \\ 
  & distant from        & \mytexttilde0.00  &  \\ 
  & remote from         & \mytexttilde0.00  &  \\ 
\midrule

% ------------------------ Group: inside\_within (5 keywords) ------------------------
\multirow{7}{*}{inside\_within} 
  & inside              & 0.56  & 7.97 & 2.74 \\ 
  & within              & 0.15  &  \\ 
  & inside of           & 0.04  &  \\ 
  & contained in        & \mytexttilde0.00  &  \\ 
  & enclosed by         & \mytexttilde0.00  &  \\ 
  & in                  & 7.22  &  \\ 
\midrule

\multirow{6}{*}{outside} 
  & outside            & 0.09  & 0.17 & 0.06 \\ 
  & out of             & 0.09  &  \\ 
  & outer              & \mytexttilde0.00  &  \\ 
  & outside of         & \mytexttilde0.00  &  \\ 
  & outlying           & \mytexttilde0.00  &  \\ 
\midrule

\multirow{10}{*}{next\_to\_beside\_adjacent} 
  & next to            & 1.79  & 3.60 & 1.24 \\ 
  & beside             & 0.62  &  \\ 
  % & alongside          & 0.15  &  \\ 
  & adjacent           & 0.34  &  \\ 
  & adjacent to        & 0.34  &  \\ 
  & by                 & 0.19  &  \\ 
  & at the side of     & 0.30  &  \\ 
  & by the side of     & \mytexttilde0.00  &  \\ 
  & side by side with  & \mytexttilde0.00  &  \\ 
  & contiguous with    & \mytexttilde0.00  &  \\ 
\midrule

\multirow{6}{*}{opposite} 
  & opposite           & 0.09  & 0.17 & 0.06 \\ 
  & opposite to        & 0.09  &  \\ 
  & opposite side of   & \mytexttilde0.00  &  \\ 
  & diagonally across  & \mytexttilde0.00  &  \\ 
  & opposite from      & \mytexttilde0.00  &  \\ 
  & opposed to         & \mytexttilde0.00  &  \\ 
\midrule

\multirow{5}{*}{facing} 
  & facing             & 0.62  & 0.62 & 0.21 \\ 
  & facing toward      & \mytexttilde0.00  &  \\ 
  & looking at         & \mytexttilde0.00  &  \\ 
  & confronting        & \mytexttilde0.00  &  \\ 
  & in view of         & \mytexttilde0.00  &  \\ 
\midrule

\multirow{4}{*}{parallel\_to} 
  & parallel to         & 0.13  & 0.13 & 0.04 \\ 
  & in line with        & \mytexttilde0.00  &  \\ 
  & aligned with        & \mytexttilde0.00  &  \\ 
  & running parallel to & \mytexttilde0.00  &  \\ 
\midrule

\newpage

\multirow{4}{*}{perpendicular\_to} 
  & perpendicular to    & 0.15  & 0.15 & 0.05 \\ 
  & perpendicular with  & \mytexttilde0.00  &  \\ 
  & orthogonal to       & \mytexttilde0.00  &  \\ 
  & at right angles to  & \mytexttilde0.00  &  \\ 
\midrule

\multirow{7}{*}{toward\_towards} 
  & toward              & 0.15  & 0.15 & 0.05 \\ 
  & towards             & \mytexttilde0.00  &  \\ 
  & proceeding to       & \mytexttilde0.00  &  \\ 
  & progressing toward  & \mytexttilde0.00  &  \\ 
  & moving toward       & \mytexttilde0.00  &  \\ 
  & heading toward      & \mytexttilde0.00  &  \\ 
  & approaching         & \mytexttilde0.00  &  \\ 
\midrule

\multirow{7}{*}{away\_from} 
  & away                & 1.40  & 2.80 \\ 
  & away from           & 1.40  &  \\ 
  & moving away from    & \mytexttilde0.00  &  \\ 
  & departing from      & \mytexttilde0.00  &  \\ 
  & receding from       & \mytexttilde0.00  &  \\ 
  & withdrawing from    & \mytexttilde0.00  &  \\ 
  & retreating from     & \mytexttilde0.00  &  \\ 
\midrule

\multirow{6}{*}{between} 
  & between         & 0.02  & 0.02 \\ 
  & among           & \mytexttilde0.00  &  \\ 
  & amid            & \mytexttilde0.00  &  \\ 
  & amidst          & \mytexttilde0.00  &  \\ 
  & amongst         & \mytexttilde0.00  &  \\ 
  & betwixt         & \mytexttilde0.00  &  \\ 
\midrule

\multirow{7}{*}{through} 
  & through          & 0.02  & 0.04 \\ 
  & passing through  & \mytexttilde0.00  &  \\ 
  & traversing       & \mytexttilde0.00  &  \\ 
  & transiting       & \mytexttilde0.00  &  \\ 
  & running through  & \mytexttilde0.00  &  \\ 
  & crossing         & 0.02  &  \\ 
  & piercing         & \mytexttilde0.00  &  \\ 
\midrule

\multirow{9}{*}{around} 
  & around         & 0.17  & 0.22 \\ 
  & circling       & \mytexttilde0.00  &  \\ 
  & encircling     & \mytexttilde0.00  &  \\ 
  & surrounding    & 0.04  &  \\ 
  & enveloped by   & \mytexttilde0.00  &  \\ 
  & enclosing      & \mytexttilde0.00  &  \\ 
  & skirting       & \mytexttilde0.00  &  \\ 
  & encompassing   & \mytexttilde0.00  &  \\ 
  & encircled by   & \mytexttilde0.00  &  \\ 
\midrule

\multirow{8}{*}{overlapping\_intersecting} 
  & overlapping          & \mytexttilde0.00  & \mytexttilde0.00 \\ 
  & overlapping with     & \mytexttilde0.00  &  \\ 
  & intersecting with    & \mytexttilde0.00  &  \\ 
  & interlacing with     & \mytexttilde0.00  &  \\ 
  & intertwined with     & \mytexttilde0.00  &  \\ 
  & interlocking         & \mytexttilde0.00  &  \\ 
  & crisscrossing        & \mytexttilde0.00  &  \\ 
  & interlaced with      & \mytexttilde0.00  &  \\ 
\midrule

\multirow{9}{*}{connected\_attached} 
  & connected to      & \mytexttilde0.00  & \mytexttilde0.00 \\ 
  & connected with    & \mytexttilde0.00  &  \\ 
  & attached to       & \mytexttilde0.00  &  \\ 
  & attached with     & \mytexttilde0.00  &  \\ 
  & linked to         & \mytexttilde0.00  &  \\ 
  & joined to         & \mytexttilde0.00  &  \\ 
  & contiguous with   & \mytexttilde0.00  &  \\ 
  & linked with       & \mytexttilde0.00  &  \\ 
  & adjoined to       & \mytexttilde0.00  &  \\ 
\midrule

\newpage

\multirow{6}{*}{within\_boundary} 
  & at the edge of     & 0.65  & 0.65 \\ 
  & at the corner of   & \mytexttilde0.00  &  \\ 
  & on the edge of     & \mytexttilde0.00  &  \\ 
  & bordering         & \mytexttilde0.00  &  \\ 
  & edged by          & \mytexttilde0.00  &  \\ 
  & at the boundary of & \mytexttilde0.00  &  \\ 
\midrule

\multirow{8}{*}{cardinal\_directions} 
  & north of       & \mytexttilde0.00  & \mytexttilde0.00 \\ 
  & south of       & \mytexttilde0.00  &  \\ 
  & east of        & \mytexttilde0.00  &  \\ 
  & west of        & \mytexttilde0.00  &  \\ 
  & northeast of   & \mytexttilde0.00  &  \\ 
  & northwest of   & \mytexttilde0.00  &  \\ 
  & southeast of   & \mytexttilde0.00  &  \\ 
  & southwest of   & \mytexttilde0.00  &  \\ 
\midrule

\multirow{7}{*}{central\_position} 
  & center of          & 0.04  & 0.60 \\ 
  & at the center of   & \mytexttilde0.00  &  \\ 
  & in the center of   & 0.04  &  \\ 
  & middle of          & 0.26  &  \\ 
  & in the middle of   & 0.26  &  \\ 
  & in the midst of    & \mytexttilde0.00  &  \\ 
  & amidst             & \mytexttilde0.00  &  \\ 
\midrule

\multirow{9}{*}{part\_of} 
  & part of          & 0.34  & 0.34 \\ 
  & has as a part    & \mytexttilde0.00  &  \\ 
  & consists of      & \mytexttilde0.00  &  \\ 
  & comprising       & \mytexttilde0.00  &  \\ 
  & including        & \mytexttilde0.00  &  \\ 
  & possessing       & \mytexttilde0.00  &  \\ 
  & containing       & \mytexttilde0.00  &  \\ 
  & consisting of    & \mytexttilde0.00  &  \\ 
  & made up of       & \mytexttilde0.00  &  \\ 
\midrule

\multirow{6}{*}{relative\_to} 
  & relative to       & 0.02  & 0.02 \\ 
  & relationship to   & \mytexttilde0.00  &  \\ 
  & in relation to    & \mytexttilde0.00  &  \\ 
  & with respect to   & \mytexttilde0.00  &  \\ 
  & regarding         & \mytexttilde0.00  &  \\ 
  & respecting        & \mytexttilde0.00  &  \\ 
\midrule

\multirow{6}{*}{movement\_along} 
  & along             & \mytexttilde0.00  & 0.15 \\ 
  & alongside         & 0.15  &  \\ 
  & running along     & \mytexttilde0.00  &  \\ 
  & stretching across & \mytexttilde0.00  &  \\ 
  & progressing along & \mytexttilde0.00  &  \\ 
  & moving along      & \mytexttilde0.00  &  \\ 
\midrule
\midrule

Total & & 100.00 & 100.00 \\
  
\end{longtable}

\end{document}